# Combining Multiple Feature Extraction Techniques for Handwritten Devnagari Character Recognition

Sandhya Arora
Department of CSE & IT
Meghnad Saha Institute of Technology
Kolkata-700107
sandhyabhagat@yahoo.com

Debotosh Bhattacharjee, Mita Nasipuri,
Dipak Kumar Basu[*], Mahantapas Kundu
Department of Computer Science and Engineering,
Jadavpur University, Kolkata, 700032, India
[*]AICTE Emeritus Fellow

*Abstract*— In this paper we present an OCR for Handwritten Devnagari Characters. Basic symbols are recognized by neural classifier. We have used four feature extraction techniques namely, intersection, shadow feature, chain code histogram and straight line fitting features. Shadow features are computed globally for character image while intersection features, chain code histogram features and line fitting features are computed by dividing the character image into different segments. Weighted majority voting technique is used for combining the classification decision obtained from four Multi Layer Perceptron(MLP) based classifier. On experimentation with a dataset of 4900 samples the overall recognition rate observed is 92.80% as we considered top five choices results. This method is compared with other recent methods for Handwritten Devnagari Character Recognition and it has been observed that this approach has better success rate than other methods.

## I. INTRODUCTION

Although there are many scripts and languages in India but not much research is done for the recognition of handwritten Indian characters. There are many pieces of work reported towards handwritten recognition of Roman, Japanese, Chinese and Arabic scripts. In this paper, we propose a system for the recognition of off-line handwritten Devnagari characters.

Devnagari is third most widely used script, used for several major languages such as Hindi, Sanskrit, Marathi and Nepali, and is used by more than 500 million people. Unconstrained Devnagari writing is more complex than English cursive due to the possible variations in the order, number, direction and shape of the constituent strokes. Devnagari script has 50 characters which can be written as individual symbols in a word (some shown in figure 1). Devnagari Character recognition is complicated by presence of multiple loops, conjuncts, upper and lower modifiers and the number of disconnected and multistroke characters [19], in a word where all characters are connected through header line. Modifiers make Optical Character recognition (OCR) with Devnagari script very challenging. OCR is further complicated by compound characters that make character separation and identification very difficult.

Although first research report on handwritten Devnagari characters was published in 1977 [8] but not much research work is done after that. At present researchers have started to work on handwritten Devnagari characters and few research reports are published recently. Hanmandlu and Murthy [9, 21] proposed a Fuzzy model based recognition of handwritten Hindi numerals and characters and they obtained 92.67% accuracy for Handwritten Devnagari numerals and 90.65% accuracy for Handwritten Devnagari characters. Bajaj et al [10] employed three different kinds of features namely, density features, moment features and descriptive component features for classification of Devnagari Numerals. They proposed multi-classifier connectionist architecture for increasing the recognition reliability and they obtained 89.6% accuracy for handwritten Devnagari numerals. Kumar and Singh [11] proposed a Zernike moment feature based approach for Devnagari handwritten character recognition. They used an artificial neural network for classification. Sethi and Chatterjee [12] proposed a decision tree based approach for recognition of constrained hand printed Devnagari characters using primitive features. Bhattacharya et al [13] proposed a Multi-Layer Perceptron (MLP) neural network based classification approach for the recognition of Devnagari handwritten numerals and obtained 91.28% results. N. Sharma and U. Pal [5] proposed a directional chain code features based quadratic classifier and obtained 80.36% accuracy for handwritten Devnagari characters and 98.86% accuracy for handwritten Devnagari numerals. In most of the works reported above, multiple classifier combination has not been reported for handwritten Devnagari characters. Most of them are based on single classifier or reported for handwritten Devnagari numerals. In this paper we present the results of multiple classifier combination for offline handwritten Devnagari character recognition.

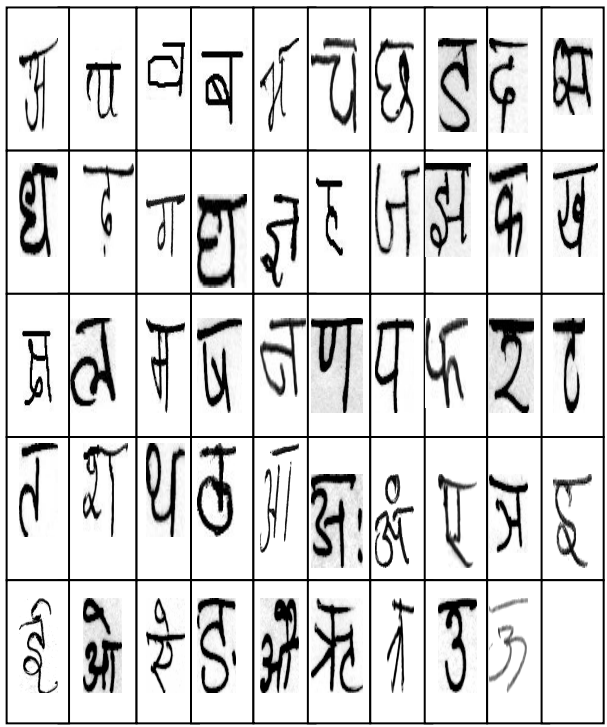

Figure 1.Some samples of Devnagari characters from the database

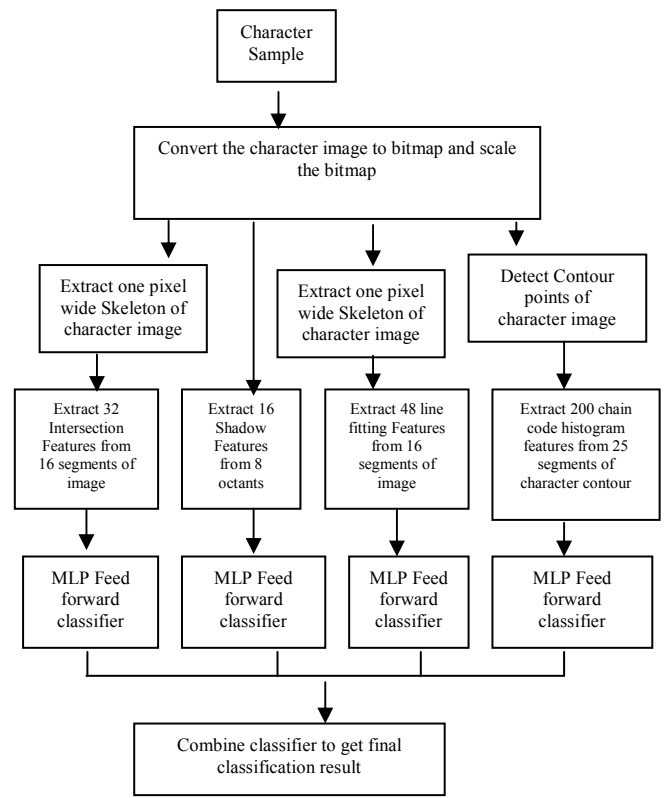

Figure 2. Block diagram to represent the proposed technique

This character recognition technique uses features obtained from Shadow, intersection and chain code histogram. Basic methodology, preprocessing, feature extraction is discussed in section 2.The recognition classifier and its combination is discussed in section 3. Results obtained after application of the technique on handwritten Devnagari characters are shown in section 4.

## II. PROPOSED METHOD

In our proposed method, we are first performing scaling of character bitmap and after that we are extracting three different features. First, 32 intersection features are extracted after performing thinning, generating one pixel wide skeleton of character image and segmenting the image into 16 segments. Second, 16 shadow features are extracted from eight octants of the character image. Third, 200 chain code histogram features are obtained by first detecting the contour points of original scaled character image, and dividing the contour image into 25 segments. For each segment chain code histogram features are obtained.

### A.. Conversion of Handwritten Character to Bitmapped Binary Images

In images, background is not always of same contrast value. For this, we designed the Dynamic Threshold Value Identification algorithm. The goal of this algorithm is to distinguish between image pixels that belong to text and those that belong to the background. The algorithm mentioned below is applied on gray scale images:-

a) Take the threshold to be 128(Mid value of 0 to 255).

b) Take all the pixels with grayscale value above 128 as background and all those with value below 128 as foreground.

c) Find the mean grayscale values of background pixels and foreground pixels.

d) Find the average of both mean values and make this the new threshold.

e) Go back to step (b) and continue this process of refining the threshold till the change of the threshold from one iteration to next becomes less than 2%.

### B. Scaling of the binary character images

Each character image is first enclosed in a tight fit rectangular boundary. The portion of the image outside this boundary is discarded. The selected portion of the character image is then scaled to the size 100 × 100 pixel using affine transformation [16]. For smoothing the contours of the character images and also for filling in some holes, we perform some morphological operations like closing and dilation on the character image.

*C. Feature Extraction*

In the following we give a brief description of the three feature sets used in our proposed multiple classifier system. Shadow features are extracted from scaled bitmapped character image. Chain code histogram features are extracted by chain coding the contour points of the scaled character bitmapped image. Intersection features are extracted from scaled, thinned one pixel wide skeleton of character image.

*Shadow Features of character--* For computing shadow features [20], the rectangular boundary enclosing the character image is divided into eight octants, for each octant shadow of character segment is computed on two perpendicular sides so a total of 16 shadow features are obtained. Shadow is basically the length of the projection on the sides as shown in figure 3. These features are computed on scaled image.

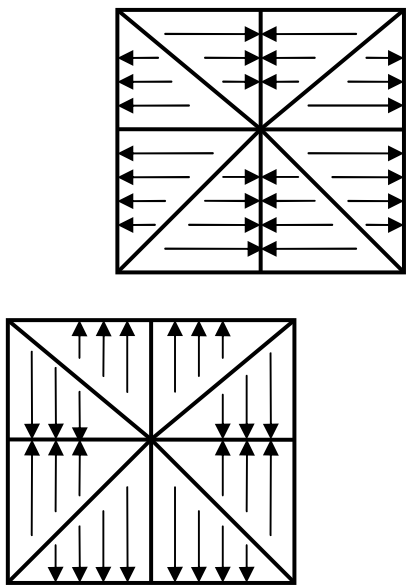

Figure 3. Shadow features

*Chain Code Histogram of Character Contour--* Given a scaled binary image, we first find the contour points of the character image. We consider a 3 × 3 window surrounded by the object points of the image. If any of the 4-connected neighbor points is a background point then the object point (P), as shown in figure 4 is considered as contour point.

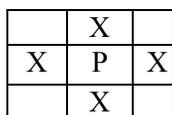

Figure 4. Contour point detection

The contour following procedure uses a contour representation called "chain coding" that is used for contour following proposed by Freeman [22], shown in figure 5a. Each pixel of the contour is assigned a different code that indicates the direction of the next pixel that belongs to the contour in some given direction. Chain code provides the points in relative position to one another, independent of the coordinate system. In this methodology of using a chain coding of connecting neighboring contour pixels, the points and the outline coding are captured. Contour following procedure may proceed in clockwise or in counter clockwise direction. Here, we have chosen to proceed in a clockwise direction.

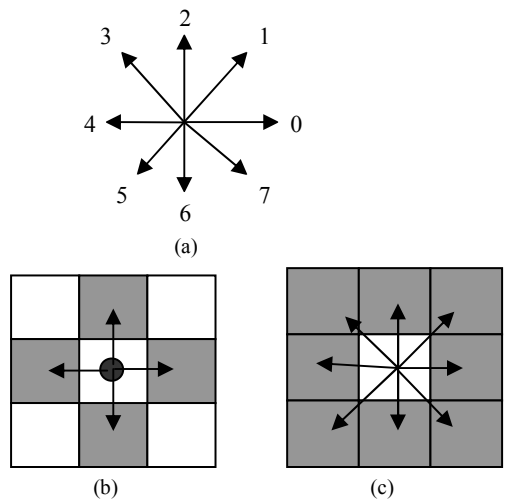

Figure 5. Chain Coding: (a) direction of connectivity, (b) 4-connectivity, (c) 8-connectivity. Generate the chain code by detecting the direction of the next-in-line pixel

The chain code for the character contour will yield a smooth, unbroken curve as it grows along the perimeter of the character and completely encompasses the character. When there is multiple connectivity in the character, then there can be multiple chain codes to represent the contour of the character. We chose to move with minimum chain code number first.

We divide the contour image in 5 × 5 blocks. In each of these blocks, the frequency of the direction code is computed and a histogram of chain code is prepared for each block. Thus for 5 × 5 blocks we get 5 × 5 × 8 = 200 features for recognition.

*Finding Intersection/Junctions in character--* An intersection, also referred to as a junction, is a location where the chain code goes in more than a single direction in an 8-connected neighborhood (Fig-6). Thinned and scaled character image is divided into 16 segments each of size 25 × 25 pixels wide. For each segment the number of open end points and junctions are calculated. Intersection point is defined as a pixel point which has more than two neighboring pixels in 8-connectivity while an open end has exactly one neighbor pixel. Intersection points are unique for a character in different segment. Thus the number of 32 features within the 16 constituent segments of the character image are collected, out of which first 16 features represents the number of open ends and rest 16 features represents number of junction points within a segment. These features are observed after image thinning and scaling as without thinning of the character image there will be multiple open end points and multiple junction points within a segment. For thinning standard algorithm [15] is used. This algorithm does not generate one-pixel wide skeleton character image and results in redundant pixels so we designed certain masks [17] so that after application of these masks some image pixels are removed so finally we get these entire characters one pixel wide. Thinned character images are shown in figure-7. Scaling is done as we are dividing the character image into fixed size segments which is not possible without fixing image size.

| P2 | P1 | P8 |
| --- | --- | --- |
| P3 | P | P7 |
| P4 | P5 | P6 |

Figure 6. 8-neighborhoods

Input Characters

Thinned Characters

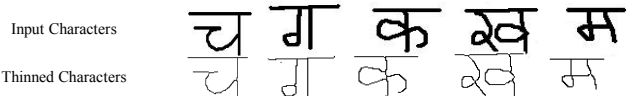

Figure 7. Results of thinning applied on isolated characters

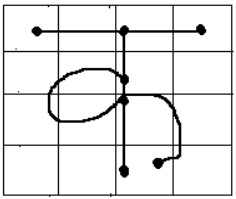

Figure 8. Intersection points and open end points

Intersection feature calculated for this character image shown in figure-8, is 1001 0000 0000 0020 0010 0010 0010 0000, where 1001 0000 0000 0020 is the information of open end points and 0010 0010 0010 0000 is the information of junction points in each segment.

*Straight Lines Fitting for Character*—A straight line $y=a_i+b_i x$ is uniquely defined by two parameters: the slope $b_i$ and the intercept $a_i$. Given a thinned and scaled binary image, we segmented it into 16 segments. For each segment image a straight line was fitted using LMS method of line fitting and calculated $a_i$ and $b_i$ as follows :-

$a_i = ( \sum x_i y_i * \sum x_i - \sum x_i^2 * \sum y_i ) / ( \sum x_i * \sum x_i - N * \sum x_i^2 )$

$b_i = ( \sum x_i y_i * N - \sum x_i * \sum y_i ) / ( N * \sum x_i^2 - \sum x_i * \sum x_i )$

Intercept feature can be used directly but using slope feature directly has a drawback related to the continuity of the representation. Straight line with slopes approaching +. And -.

Has a very similar orientation but obviously, would be represented with extremely different values, so we used two features f1 and f2 based on slope, where :

$f1 = 2 * b_i / ( 1 + b_i^2 )$

$f2 = ( 1 - b_i^2 ) / ( 1 + b_i^2 )$

So total calculated line fitting based features are 48 for a 16 segmented image out of which 16 are $a_i$ values, 16 are f1 values and 16 are f2 values.

### III. DEVNAGARI CHARACTER RECOGNITION

We used the same MLP with 3 layers including one hidden layer for four different feature sets consisting of 32 intersection features, 16 shadow features, 48 line fitting based features and 200 chain code histogram features. The experimental results obtained while using these features for recognition of handwritten Devnagari characters is presented in the next section. At this stage all characters are non-compound, single characters so no segmentation is required.

The classifier is trained with standard Backpropagation [14]. It minimizes the sum of squared errors for the training samples by conducting a gradient descent search in the weight space. As activation function we used sigmoid function. Learning rate and momentum term are set to 0.8 and 0.7 respectively. As activation function we used the sigmoid function. Numbers of neurons in input layer of MLPs are 32, 16, 48 or 200, for intersection features, shadow features, line fitting features and chain code histogram features respectively. Number of neurons in Hidden layer is not fixed, we experimented on the values between 20-70 to get optimal result and finally it was set to 20, 30, 40 and 70 for intersection features, shadow features, line fitting features and chain code histogram features respectively. The output layer contained one node for each class., so the number of neurons in output layer is 49. And classification was accomplished by a simple maximum response strategy.

#### A. Classifier Combination

The ultimate goal of designing pattern recognition system is to achieve the best possible classification performance. This objective traditionally led to the development of different classification scheme for any pattern recognition problem to be solved. The result of an experimental assessment to the different design would then be the basis for choosing one of the classifiers as the final solution to the problem. It had been observed in such design studies, that although one of the designs would yield the best performance, the sets of patterns misclassified by the different classifiers would not necessarily overlap. This suggested that different classifier designs potentially offered complementary information about the pattern to be classified which could be harnessed to improve the performance of the selected classifier. So instead of relying on a single decision making scheme we can combine classifiers.

Combination of individual classifier outputs overcomes deficiencies of features and trainability of single classifiers. Outputs from several classifiers can be combined to produce a more accurate result. Classifier combination takes two forms: combination of like classifiers trained on different data sets and combination of dissimilar classifiers. We have four similar Neural networks classifiers as discussed above, which are trained on 32 intersection features, 16 shadow features, 48 line fitting features and 200 chain code features respectively. The outputs are confidences associated with each class. As these outputs can not be compared directly, we used an aggregation function for combining the results of all four classifiers. Our strategy is based on weighted majority voting scheme as described below.

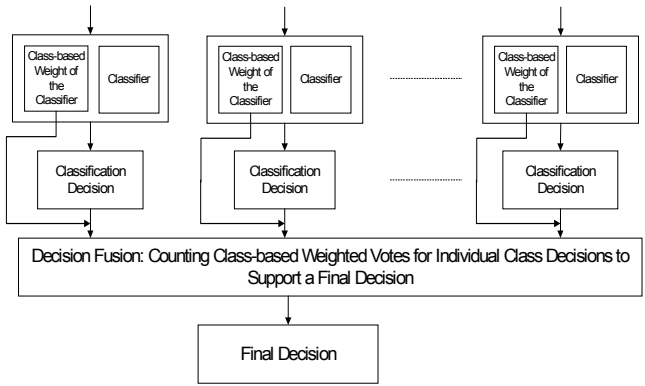

Figure 9. Weighted majority voting technique for combining classifiers

So if $k^{th}$ classifier decision to assign the unknown pattern to the $i^{th}$ class is denoted by $O_{ik}$ with $1 \leq i \leq m$, $m$ being the number of classes, then the final combined decision $d_i^{com}$ supporting assignment to the $i^{th}$ class takes the form of :-

$$d_i^{com} = \sum_{k=1,2,3,4} \omega_k * O_{ik} \quad ........ 1 \leq i \leq m$$

The final decision $d^{com}$ is therefore :-

$$d^{com} = \max_{1 \leq i \leq m} d_i^{com}$$

$$\omega_k = \frac{d_k}{\sum_{k=1}^{4} d_k}$$

where $m = 50$ and $\omega_1, \omega_2, \omega_3$ are 0.349, 0.197, 0.326 and 0.128 respectively as $d_1 > d_3 > d_2 > d_4$

$d_1$=64.90% result of classifier trained with chaincode histogram features

$d_2$=36.71% result of classifier trained with Intersection features

$d_3$= 60.59% result of classifier trained with Shadow features.

$d_4$=24.83% result of classifier trained with straight line fitting features

## IV. RESULTS

The handwritten Devnagari character dataset is collected from CVPR Unit, ISI, kolkata. For preparation of the training and test sets a sample set of 4900 handwritten devnagari characters are considered. A training set of 3332 samples and test set of 1568 samples are formed. Results are shown in table 2. We have not considered devnagari numerals right now. We considered top 1 choice, top 2 choices, top 3 choices, top 4 choices and top 5 choices for classification results. We applied 3-fold cross validation testing. We divided the whole dataset into three parts. In first fold, first two parts are used for training and third part is used for testing. In second fold, first and third part is used for training and second part is used for testing. In fold three, second and third part is used for training and first part is used for testing. We used the same MLP for three different feature sets as discussed above in section 3. The results of three MLPs are shown in table 1.

Table I. Results of three different MLP

| MLP | Input layer Neuron | Hidden Layer Neuron | Output Layer Neuron | Result |
|---|---|---|---|---|
| Line Fitting based | 48 | 40 | 49 | 24.83% |
| Intersection Features based | 32 | 20 | 49 | 36.71% |
| Shadow Feature based | 16 | 30 | 49 | 60.59% |
| Chain Code Histogram Feature based | 200 | 70 | 49 | 64.90% |

Table II. Top Choices Results

| S. No. | Proposed method result | Accuracy obtained |
|---|---|---|
| 1 | Top 1 choice | 70.39% |
| 2 | Top 2 choices | 81.44% |
| 3 | Top 3 choices | 86.21% |
| 4 | Top 4 choices | 88.95% |
| 5 | Top 5 choices | 92.80% |

Table III.. Recognition success

| S. No. | Method proposed by | Accuracy obtained |
|---|---|---|
| 1 | Kumar and Singh [4] | 80% |
| 2 | N. Sharma, U. Pal, F. Kimura, and S. Pal [5] | 80.36% |
| 3 | M. Hanmandlu, O.V. R. Murthy, V.K. Madasu[21] | 90.65% |
| 4 | Proposed method (considering top 5 choices) | 92.80% |

We used another classification strategy where we have used the four same above said classifiers, and character is said to be classified in the class for which it is giving maximum response for any one of the four classifiers. So the character is said to be classified if it has been classified by any one of the classifier. It was giving 80.71% accuracy, which is better than results reported by Kumar, singh[4] and N. Sharma, U. Pal, F. Kimura, S. Pal [5]. We used 4900 samples dataset which is larger than 4750 used by M. Hanmandlu, O.V. R. Murthy, V.K. Madasu[21]. They reported 69.78% overall recognition rate for handwritten devnagari characters which is similar to our top 1 choices result. Their coarse classification is giving 90.65% accuracy but we have obtained 92.80% accuracy.

## V. CONCLUSION

This paper proposes an off-line handwritten Devnagari character recognition system consisting of three different features using simple feedforward Multilayer Perceptrons. We obtained encouraging results. This technique can be applied on handwritten Devnagari numerals also and it will also be helpful for research towards other similar Indian scripts like Tamil, Bengali, Gujarati, Gurmukhi, Oriya.

## ACKNOWLEDGMENT

Authors are thankful to the "Centre for Microprocessor Application for Training Education and Research" and "Project on Storage Retrieval and Understanding of Video for Multimedia", at the Department of Computer Science and Engineering, Jadavpur University, Kolkata-700032 for providing the necessary facilities for carrying out this work. Authors are also thankful to the CVPR Unit, ISI Kolkata for providing the dataset of Handwritten Devnagari Characters. First author gratefully acknowledge the support of the Meghnad Saha Institute of Technology for carrying out this research work.